%% file: main.tex
\documentclass[10pt,twocolumn,letterpaper]{article}

\PassOptionsToPackage{dvipsnames,table,xcdraw}{xcolor}
\usepackage{xcolor}

\usepackage{simpleConference}
\usepackage{times}
\usepackage{graphicx}
\usepackage{amssymb}
\usepackage{hyperref}
\hypersetup{colorlinks, allcolors=black}

\usepackage{booktabs}
\usepackage{subcaption}
\usepackage{multirow}
\usepackage{tikz}
\usepackage{amsmath}
\usetikzlibrary{positioning, fit, shapes.geometric, arrows, calc}

\usepackage{pifont}
\usepackage{bm} 
\usepackage{pgfplots}
\usepackage{tikz}
\usetikzlibrary{shapes.geometric}
\usepackage{listings}
\begin{document}
\title{BARIS: Boundary-Aware Refinement with Environmental Degradation Priors
for Robust Underwater Instance Segmentation}

\author{Pin-Chi Pan\textsuperscript{*}, Soo-Chang Pei\textsuperscript{\dag} \\ \\
\textsuperscript{*} Graduate Institute of Communication Engineering, National Taiwan University \\
\textsuperscript{\dag} Department of Electrical Engineering,
National Taiwan University
}

\maketitle
\thispagestyle{empty}
\input{sec/0_abstract}
\input{sec/1_intro}
\input{sec/2_relatedwork}

\input{sec/3_method}

\input{sec/4_experiments}

\input{sec/5_results}

\input{sec/6_conclusion}

\bibliographystyle{abbrv}
\bibliography{main}

\input{sec/X_suppl}

\end{document}

%% file: sec/0_abstract.tex
\begin{abstract} 
Underwater instance segmentation is challenging due to adverse visual conditions such as light attenuation, scattering, and color distortion, which degrade model performance. In this work, we propose \textbf{BARIS-Decoder} (\textbf{B}oundary-\textbf{A}ware \textbf{R}efinement Decoder for \textbf{I}nstance \textbf{S}egmentation), a framework that enhances segmentation accuracy through feature refinement. To address underwater degradations, we introduce the \textbf{Environmental Robust Adapter (ERA)}, which efficiently models underwater degradation patterns while reducing trainable parameters by over 90\% compared to full fine-tuning. The integration of BARIS-Decoder with ERA-tuning, referred to as \textbf{BARIS-ERA}, achieves state-of-the-art performance, surpassing Mask R-CNN by 3.4 mAP with a Swin-B backbone and 3.8 mAP with ConvNeXt V2. Our findings demonstrate the effectiveness of BARIS-ERA in advancing underwater instance segmentation, providing a robust and efficient solution. 
\vspace{8pt}
\end{abstract}

%% file: sec/1_intro.tex
\vspace{-8pt}
\section{Introduction}
\label{sec:intro}
\vspace{-8pt}
\begin{figure}[t]
    \centering
    \hspace*{0.2cm} 
    \begin{minipage}{1.0\linewidth} 
        \centering
        \begin{subfigure}{0.23\linewidth}
            \centering
            \includegraphics[width=\linewidth]{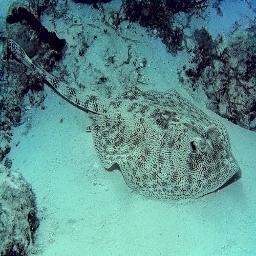}
            \caption{Input}
        \end{subfigure}
        \begin{subfigure}{0.23\linewidth}
            \centering
            \includegraphics[width=\linewidth]{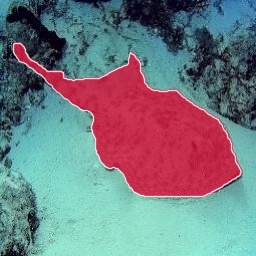}
            \caption{WaterMask}
        \end{subfigure}
        \begin{subfigure}{0.23\linewidth}
            \centering
            \includegraphics[width=\linewidth]{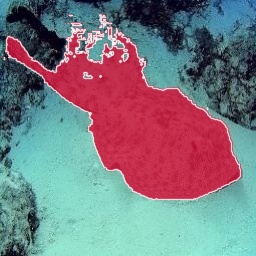}
            \caption{USIS-SAM}
        \end{subfigure}
        \begin{subfigure}{0.23\linewidth}
            \centering
            \includegraphics[width=\linewidth]{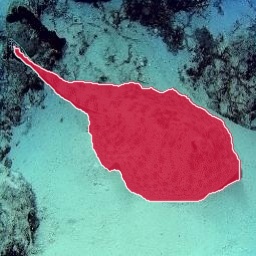}
            \caption{\textbf{Ours}}
        \end{subfigure}
    \end{minipage}
    
    \hspace{-0.5cm}
    \begin{tikzpicture}
        \begin{axis}[
            ybar,
            bar width=0.46cm, 
            width=1.1\linewidth, height=0.7\linewidth, 
            symbolic x coords={Mask R-CNN, WaterMask, USIS-SAM, BARIS-ERA (Ours)},
            xtick=data,
            xticklabels={
                \raisebox{12pt}{Mask R-CNN},
                \raisebox{12pt}{WaterMask},
                \raisebox{12pt}{USIS-SAM},
                \shortstack{\textbf{BARIS-ERA} \\ \textbf{(Ours)}}
            },
            xticklabel style={xshift=10pt, yshift=-10pt, rotate=20, anchor=east, font=\fontsize{10pt}{7pt}},
            xtick pos=left,
            ytick style={draw=none},
            ymin=15, ymax=62,
            ylabel={AP(\%)},
            ylabel style={yshift=-12pt, font=\fontsize{7pt}{7pt}},
            enlarge x limits=0.2, 
            legend style={
                at={(0.31, 0.84)}, 
                anchor=south, 
                legend columns=-1, 
                column sep=4pt, 
                font=\fontsize{7pt}{7pt},
            },
            legend cell align={left},
        ]

        \definecolor{mapcolor}{RGB}{31, 119, 180}  
        \definecolor{ap50color}{RGB}{255, 127, 14} 
        \definecolor{ap75color}{RGB}{44, 160, 44}  

        \addplot[fill=mapcolor] coordinates {(Mask R-CNN, 28.2) (WaterMask, 30.1) (USIS-SAM, 29.4) (BARIS-ERA (Ours), 31.6)};
        \addplot[fill=ap75color] coordinates {(Mask R-CNN, 32.1) (WaterMask, 33.5) (USIS-SAM, 32.3) (BARIS-ERA (Ours), 33.6)};
        \addplot[fill=ap50color] coordinates {(Mask R-CNN, 46.6) (WaterMask, 49.0) (USIS-SAM, 45.0) (BARIS-ERA (Ours), 52.0)};

        \node[above, xshift=-16pt, yshift=1pt, font=\small, inner sep=1pt] at (axis cs:{Mask R-CNN}, 28.2) {28.2};
        \node[above, xshift=-1pt, yshift=1pt, font=\small, inner sep=1pt] at (axis cs:{Mask R-CNN}, 32.1) {32.1};
        \node[above, xshift=15pt, yshift=1pt, font=\small, inner sep=1pt] at (axis cs:{Mask R-CNN}, 46.6) {46.6};

        \node[above, xshift=-16pt, yshift=1pt, font=\small, inner sep=1pt] at (axis cs:{WaterMask}, 30.1) {30.1};
        \node[above, xshift=-1pt, yshift=1pt, font=\small, inner sep=1pt] at (axis cs:{WaterMask}, 33.5) {33.5};
        \node[above, xshift=15pt, yshift=1pt, font=\small, inner sep=1pt] at (axis cs:{WaterMask}, 49.0) {49.0};

        \node[above, xshift=-16pt, yshift=1pt, font=\small, inner sep=1pt] at (axis cs:{USIS-SAM}, 29.4) {29.4};
        \node[above, xshift=-1pt, yshift=1pt, font=\small, inner sep=1pt] at (axis cs:{USIS-SAM}, 32.3) {32.3};
        \node[above, xshift=15pt, yshift=1pt, font=\small, inner sep=1pt] at (axis cs:{USIS-SAM}, 45.0) {45.0};

        \node[above, xshift=-17pt, yshift=1pt, text=red, font=\bfseries\small, inner sep=1pt] at (axis cs:{BARIS-ERA (Ours)}, 31.6) {31.6};
        \node[above, xshift=-1pt, yshift=1pt, text=red, font=\bfseries\small, inner sep=1pt] at (axis cs:{BARIS-ERA (Ours)}, 33.6) {33.6};
        \node[above, xshift=15pt, yshift=1pt, text=red, font=\bfseries\small, inner sep=1pt] at (axis cs:{BARIS-ERA (Ours)}, 52.0) {52.0};

        \legend{mAP, AP$_{75}$, AP$_{50}$}

        \end{axis}
    \end{tikzpicture}
    \vspace{-16pt}
    \caption{Comparison of our approach with state-of-the-art methods on the UIIS dataset. USIS-SAM \cite{lian2024diving} uses a ViT-H backbone, while all other methods adopt Swin-B. Our BARIS-ERA method achieves the best performance across all AP metrics.}
    \vspace{-8pt}
    \label{fig:fig1}
\end{figure}
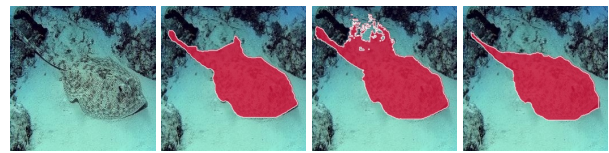
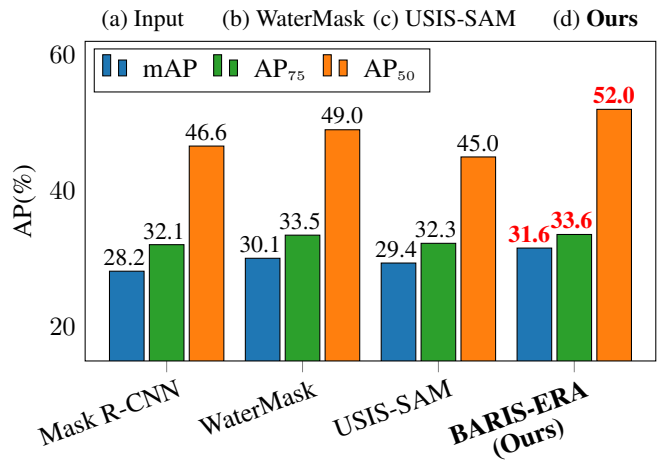
Instance segmentation is a fundamental task in computer vision, with applications in autonomous robotics, medical imaging, remote sensing, and environmental monitoring \cite{liu2020real, chen2020perceptual}. Although significant progress has been made in terrestrial settings, underwater instance segmentation remains challenging due to visual distortions, including light attenuation, scattering, and wavelength-dependent color shifts \cite{akkaynak2017space, mcglamery1980computer, jaffe1990computer}. These distortions degrade image quality, obscure object boundaries, and vary with depth and lighting. Furthermore, suspended particles (marine snow) and surface reflections complicate segmentation, causing misclassified regions and loss of details. As a result, land segmentation models often underperform in underwater datasets due to disparities in object properties and environmental conditions, such as texture, lighting, and water clarity.

Existing methods adopt primarily multi-scale feature fusion \cite{lian2023watermask} and adapter-based tuning \cite{lian2024diving} to improve segmentation accuracy. Multi-scale fusion techniques, such as RefineMask \cite{zhang2021refinemask} and WaterMask \cite{lian2023watermask}, enhance feature representation by aggregating spatial information across resolutions. However, while they improve contextual understanding, they do not explicitly refine object boundaries, leading to errors in complex scenes with densely clustered objects, such as overlapping fish schools. Adapter-based tuning methods allow pretrained models to adapt efficiently with fewer parameters. USIS-SAM \cite{lian2024diving} introduces underwater priors via adapters, enhancing environmental adaptation. However, existing adapter techniques primarily focus on feature modulation and lack explicit mechanisms for handling boundary ambiguities or directly counteracting underwater distortions like scattering and color shifts. As a result, segmentation errors persist in degraded conditions.

\begin{figure}[t]
  \centering
  \includegraphics[width=1.0\linewidth]{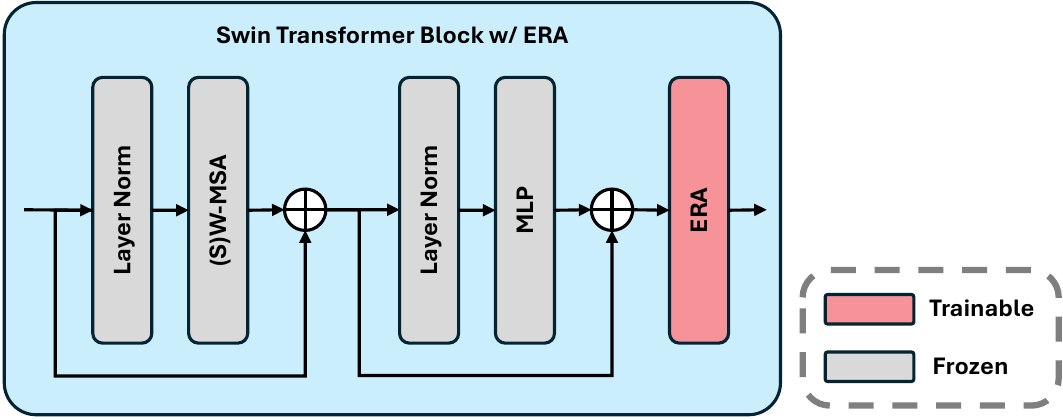}
  \caption{The Environmental Robust Adapter (ERA) integrated into a Swin Transformer block. ERA is positioned at the end of each block, while the rest of the network remains frozen. This design enables efficient adaptation to underwater distortions without modifying the core model architecture.}
  \vspace{-8pt}
  \label{fig:fig2}
\end{figure}

To address these challenges, we propose a generalizable instance segmentation framework called BARIS-Decoder. Unlike conventional feature pyramid networks, BARIS-Decoder incorporates a Multi-Stage Gated Refinement Network (MSGRN) for progressive feature refinement and Depthwise Separable Upsampling (DSU) for efficient multi-scale fusion, leading to more precise mask delineation. In addition, we introduce the Environmental Robust Adapter (ERA), an adapter-based tuning approach designed for underwater imagery. As shown in Figure~\ref{fig:fig2}, ERA is inserted at the end of each transformer or convolutional block (Swin Transformer or ConvNeXt V2), and leverages lightweight adapters to learn environmental priors, effectively counteracting underwater distortions while reducing trainable parameters by over 90\% compared to full fine-tuning. To further enhance boundary accuracy, we propose the Boundary-Aware Cross-Entropy Loss (BACE Loss), which improves mask quality by refining object boundaries. By integrating all components, BARIS-ERA dynamically adapts to underwater degradation patterns while improving segmentation robustness without adding excessive inference complexity.

Figure~\ref{fig:fig1} presents a qualitative and quantitative comparison of the BARIS-Decoder with ERA-tuning (BARIS-ERA) versus existing methods in the UIIS data set, demonstrating consistent improvements in evaluation metrics. Integrating BARIS-ERA into Mask R-CNN achieves AP gains of 3.4, 1.5, and 5.4 in mAP, AP$_{75}$, and AP$_{50}$, respectively, over the baseline Mask R-CNN \cite{he2017mask} using Swin Transformer backbones. These results highlight the effectiveness of our approach in addressing underwater imaging challenges. Our contributions are threefold. 1) We introduce BARIS-Decoder, which enhances instance segmentation by refining multi-scale feature representations and improving boundary precision for more accurate mask predictions. 2) We propose ERA, an efficient adapter-based tuning method that uses environmental priors to mitigate underwater distortions while significantly reducing trainable parameters. 3) We develop BACE Loss, a novel boundary-aware loss function that improves segmentation accuracy by refining object contours. Extensive experiments validate the effectiveness of BARIS-ERA, establishing it as a state-of-the-art approach for underwater instance segmentation.

%% file: sec/2_relatedwork.tex
\vspace{-4pt}
\section{Related Work}
\label{sec:relatedwork}
\vspace{-4pt}

\begin{figure*}[h]
    \centering
    \includegraphics[width=17.5cm]{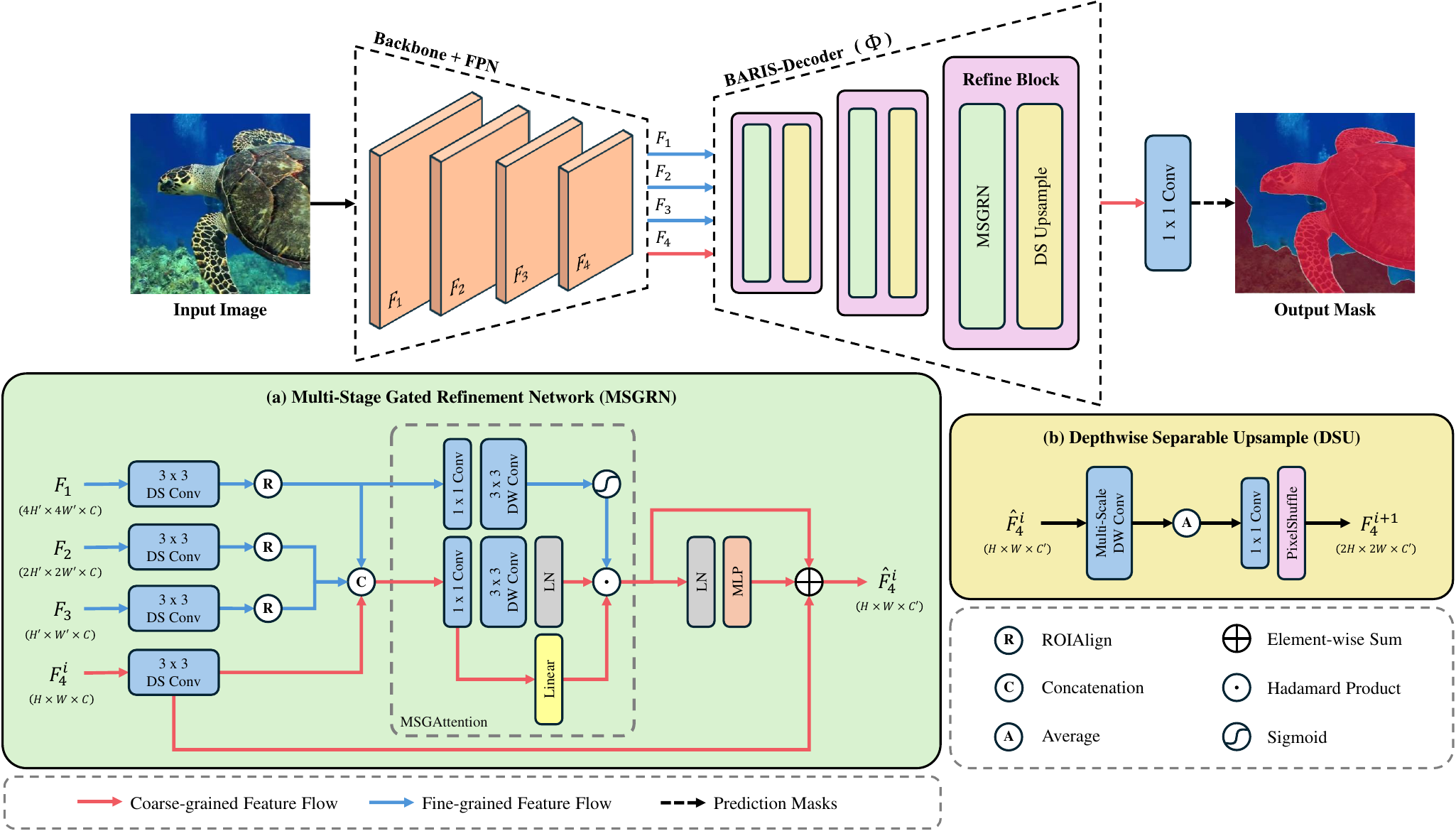}
    \caption{The architecture of the proposed BARIS-Decoder for underwater image instance segmentation. BARIS-Decoder consists of (a) Multi-Stage Gated Refinement Network (defined in Section \ref{MSGRN}) and (b) Depthwise Separable Upsample (defined in Section \ref{DSUpsample}).}
    \label{fig:fig3}
    \vspace{-8pt}
\end{figure*}

This section reviews recent advancements in underwater image segmentation and adapter-tuning techniques.

\vspace{-1pt} 
\subsection{Underwater Image Segmentation} 
\vspace{-4pt} 
Underwater image segmentation remains challenging due to environmental distortions such as light attenuation, scattering, and color degradation, which obscure object boundaries and reduce feature contrast. To address these challenges, benchmark datasets like EUVP \cite{islam2020fast}, UIEBD \cite{li2019underwater}, and SAUD \cite{jiang2022underwater} have primarily focused on image enhancement and color correction techniques. Meanwhile, datasets such as UIIS \cite{lian2023watermask} and DeepFish \cite{garcia2022efficient} emphasize biodiversity representation, promoting research in fine-grained instance segmentation.

Recent efforts have explored multi-scale feature refinement to improve segmentation accuracy in underwater conditions. WaterMask \cite{lian2023watermask} and RefineMask \cite{zhang2021refinemask} enhance object delineation by aggregating multi-scale features inspired by Feature Pyramid Networks (FPN) \cite{lin2017feature}, yet they struggle with boundary ambiguity and fine-detail preservation in low-contrast regions. USIS-SAM \cite{lian2024diving} integrates underwater priors into a vision transformer \cite{dosovitskiy2020image}-based segmentation framework using prompt-based learning, yet its reliance on high-level semantic prompts limits its ability to refine fine-grained boundary details. Although its approach is adaptable to different backbone architectures, models using large transformers, such as ViT-H, experience significantly higher computational costs and slower inference, posing challenges for real-time underwater applications like AUVs. Achieving robust segmentation across diverse underwater environments remains an open challenge, requiring a balance of multi-scale feature aggregation, environmental adaptation, and computational efficiency.

\vspace{-1pt} 
\subsection{Adapter-Tuning} 
\vspace{-4pt} 
Adapter-tuning is an efficient transfer learning technique that introduces small, trainable modules into frozen pretrained networks, reducing the need for full fine-tuning. Originally developed for natural language processing \cite{houlsby2019parameter, tinn2023fine}, this approach has gained traction in vision tasks through methods like AdaptFormer \cite{chen2022adaptformer}, Polyhistor \cite{liu2022polyhistor}, and Mona-tuning \cite{yin2023adapter}. These techniques have demonstrated success in classification and dense prediction tasks by enabling models to adapt to new domains with fewer trainable parameters. In underwater segmentation, USIS-SAM \cite{lian2024diving} incorporates adapter-based tuning to integrate domain-specific priors. However, existing adapter methods primarily focus on feature modulation and do not explicitly counteract underwater-specific distortions, such as scattering and wavelength-dependent attenuation. While adapter-tuning efficiently reduces training costs, its effectiveness in handling complex underwater degradations and segmentation challenges remains an area requiring further exploration.

%% file: sec/3_method.tex
\vspace{-4pt}
\section{Method} 
\vspace{-8pt} 
This section presents our proposed method, consisting of three main components: BARIS-Decoder (Section \ref{BARIS-Decoder}), ERA-Tuning (Section \ref{ERA}), and Boundary-Aware Cross-Entropy Loss (Section \ref{BACE_loss}).

\subsection{BARIS-Decoder}\label{BARIS-Decoder}
\vspace{-8pt}
The BARIS-Decoder (see Figure \ref{fig:fig3}) is designed to refine instance segmentation masks through multi-stage feature fusion. Unlike previous approaches that primarily rely on the final-stage feature map $\boldsymbol{F}_4$ \cite{lin2017feature, wang2019panet, wang2020deep}, BARIS-Decoder aggregates multi-scale features from all backbone stages ($\boldsymbol{F}_1$ to $\boldsymbol{F}_4$) to enhance spatial precision. While sharing a similar multi-scale refinement concept with RefineMask \cite{zhang2021refinemask}, BARIS-Decoder introduces two key innovations: 1) the Multi-Stage Gated Refinement Network (MSGRN) for progressive feature refinement and 2) the Depthwise Separable Upsample (DSU) module for efficient resolution enhancement. These components improve segmentation results while maintaining computational efficiency. The final segmentation mask $\boldsymbol{M}_{out}$ is generated as follows:
\begin{equation} 
\boldsymbol{M}_{out} = Conv_{1\times1}(\Phi(\boldsymbol{F}_1, \boldsymbol{F}_2, \boldsymbol{F}_3, \boldsymbol{F}_4)), 
\end{equation} 
where $\Phi(\cdot)$ represents the BARIS-Decoder, which processes multi-scale features using a sequence of refinement blocks. Each block applies MSGRN and DSU to iteratively improve feature quality:

\vspace{-8pt}
\begin{align}
\boldsymbol{\hat{F}}^{i}_4 &= \mathcal{M}^i_{MSGRN}(\boldsymbol{F}_1, \boldsymbol{F}_2, \boldsymbol{F}_3, \boldsymbol{F}^i_4), \notag \\
\boldsymbol{F}^{i+1}_4 &= \mathcal{M}^i_{DSU}(\boldsymbol{\hat{F}}^{i}_4),
\end{align}
where $\boldsymbol{F}^i_4$ are the features from the $i$-th refinement stage.

\vspace{-2pt}
\subsubsection{Multi-Stage Gated Refinement Network}\label{MSGRN}
\vspace{-2pt}
Multi-Stage Gated Refinement Network (MSGRN) enhances spatial details by progressively refining multi-scale features, as illustrated in Figure \ref{fig:fig3}(a). Unlike conventional fusion methods that assign equal importance to all scales, MSGRN employs Multi-Scale Gated Attention (MSGAttention) to selectively emphasize informative regions and suppress redundancy, improving boundary precision.

Inspired by High-Order Spatial Attention (HSA) from SegAdapter \cite{peng2024simple}, which modulates global features via self-gating, MSGAttention adaptively adjusts feature weights at multiple scales to refine object boundaries. The process begins with depthwise separable convolutions (DSConv) for multi-scale feature extraction:

\vspace{-8pt}
\begin{align}
\boldsymbol{X}_n &= DSConv_{3 \times 3}(\boldsymbol{F}_n),\ n \in \{1, 2, 3, 4\}, \notag \\
\boldsymbol{X}'_n &= ROIAlign(\boldsymbol{X}_n),\ n \in \{1, 2, 3\}, \notag \\
\boldsymbol{X} &= Concat(\boldsymbol{X}_4, \{\boldsymbol{X}'_n\}^{3}_{n=1})
\end{align}
MSGAttention then dynamically modulates the contribution of each scale:

\vspace{-8pt}
\begin{align}
\boldsymbol{\hat{X}} &= Conv_{1\times1}(\boldsymbol{X}), \notag \\
\boldsymbol{Y} &= LN(Conv_{3\times3}(\boldsymbol{\hat{X}})), \notag \\
\boldsymbol{V} &= Linear(\boldsymbol{\hat{X}}), \notag \\
\boldsymbol{W} &= DSConv_{3 \times 3}(\boldsymbol{X}'_1), \notag \\
\boldsymbol{\hat{Z}} &= MSGAttention(\boldsymbol{X}, \boldsymbol{X}'_1) = \sigma_{sig}(\boldsymbol{W}) \odot (\boldsymbol{Y} \odot \boldsymbol{V}), \notag \\
\boldsymbol{Z} &= FFN(\boldsymbol{\hat{Z}}) = MLP(LN(\boldsymbol{\hat{Z}})) + \boldsymbol{\hat{Z}}
\end{align}
Here, $\sigma_{sig}$ represents the sigmoid activation, $\odot$ denotes the Hadamard product, and $LN$ denotes layer normalization \cite{xu2019understanding}. While SegAdapter’s HSA globally adjusts features using high-level semantic priors, MSGAttention locally refines multi-scale features to enhance spatial details. To maintain spatial consistency, MSGRN integrates residual connections:

\vspace{-4pt}
\begin{equation}
\boldsymbol{\hat{F}}^{i}_4 = \boldsymbol{Z} + \boldsymbol{X}_4,
\end{equation}
By hierarchically refining features with selective attention, MSGRN improves segmentation accuracy while preserving spatial structure.

\vspace{-2pt}
\subsubsection{Depthwise Separable Upsample}\label{DSUpsample}
\vspace{-2pt}
The Depthwise Separable Upsample (DSU) module, shown in Figure \ref{fig:fig3}(b), enhances spatial resolution while preserving feature integrity. Unlike bilinear interpolation, DSU combines multi-scale depthwise convolutions with pixel shuffle, capturing fine-grained details efficiently. This approach enables effective multi-level feature fusion (\(\boldsymbol{F}_1\) to \(\boldsymbol{F}_4\)) while reducing computational overhead.

First, multi-scale depthwise convolutions extract high-frequency details across varying receptive fields:

\vspace{-6pt} 
\begin{equation} 
\boldsymbol{\hat{F}}^{i, j}_4 = DWConv_{j \times j}(\boldsymbol{\hat{F}}^{i}_4),\ j \in \{3, 5, 7\}.
\end{equation}

Next, the extracted features are aggregated and refined through a lightweight upsampling step:

\vspace{-12pt}
\begin{equation} 
\boldsymbol{F}^{i+1}_4 = PS(Conv_{1 \times 1}(\text{Average}(\{ \boldsymbol{\hat{F}}^{ i,j}_4 \}_{j \in \{3, 5, 7\}}))),
\end{equation}
where \( PS \) denotes the pixel shuffle operation. By leveraging pixel shuffle, DSU efficiently merges hierarchical features while maintaining spatial consistency. This design enhances segmentation accuracy without incurring the computational cost of transposed convolutions.

\begin{figure}[t]
  \centering
  \includegraphics[width=1.0\linewidth]{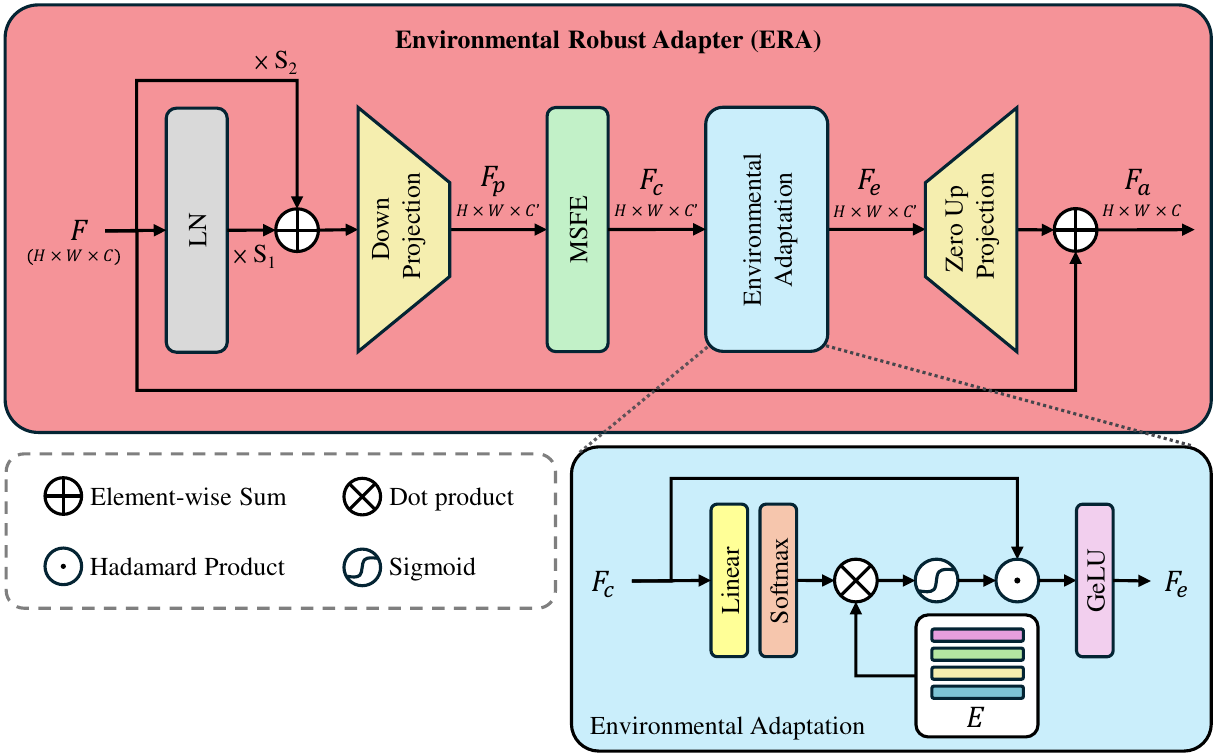}
  \caption{The architecture of the Environmental Robust Adapter (ERA). ERA enhances feature representations through multi-scale feature extraction (MSFE) and environmental adaptation.}
  \label{fig:fig4}
  \vspace{-10pt}
\end{figure}

\subsection{ERA-tuning}\label{ERA}
\vspace{-4pt}
ERA-tuning, inspired by Mona-Tuning \cite{yin2023adapter} and conceptually related to USIS-SAM \cite{lian2024diving}, employs an adapter-based strategy specifically designed for underwater degradation modeling. Unlike USIS-SAM, which integrates general semantic priors, ERA learns environmental embeddings to explicitly capture underwater degradation patterns.

As shown in Figure \ref{fig:fig4}, the input features first pass through a normalization layer, followed by two learnable scaling factors, \( S_1 \) and \( S_2 \), which adaptively modulate the feature representation. ERA then applies a down-projection, mapping features from \( \mathbb{R}^{H \times W \times C} \) to \( \mathbb{R}^{H \times W \times C'} \), where \( C' = C / \gamma \). The ratio \( \gamma \) controls feature compression and influences adaptability, as analyzed in Section \ref{Ablation}. Beyond environmental adaptation, effective feature extraction is crucial for robust segmentation. To further enhance spatial representations, we introduce the Multi-Scale Feature Extraction (MSFE) module.

\subsubsection{Multi-Scale Feature Extraction}
Multi-Scale Feature Extraction (MSFE) enhances spatial representations by capturing information at multiple receptive fields. The design is inspired by iFormer \cite{si2022inception}, where diverse kernel sizes enable robust feature learning. Specifically, MSFE applies multiple depthwise separable convolutions and max-pooling layers to improve feature discrimination:

\vspace{-8pt}
\begin{align}
\boldsymbol{F}_{s, max} &= Conv_{1\times1}(MaxPooling(\boldsymbol{F}_p)), \notag \\
\boldsymbol{F}^j_{s, conv} &= DWConv_{1\times j}(DWConv_{j\times1}(\boldsymbol{F}_p)), j \in \{3, 5, 7\}, \notag \\
\boldsymbol{F}_s = &Average(\boldsymbol{F}_{s, max}, \{\boldsymbol{F}^j_{s, conv}\}_{j \in \{3, 5, 7\}}) + \boldsymbol{F}_p, 
\end{align}
To further enhance feature representation, MSFE integrates a Channel Attention (CA) mechanism following USIS-SAM \cite{lian2024diving}. The CA module dynamically reweights feature channels to emphasize discriminative spectral information:

\vspace{-8pt}
\begin{align}
\boldsymbol{S} &= Conv_{1\times1}(\delta(Conv_{1\times1}(GAP(\boldsymbol{F}_s)))), \notag \\
\boldsymbol{F}_c &= CA(\boldsymbol{\hat{F}}_s) = \boldsymbol{\hat{F}}_s \odot \sigma_{sig}(\boldsymbol{S}),
\end{align}
where $\delta$ is the ReLU activation, $\sigma_{sig}$ is the sigmoid function, and $\odot$ denotes element-wise multiplication. By adaptively adjusting channel-wise feature importance, this mechanism strengthens critical feature representations.

\subsubsection{Environmental Adaptation}
The environmental adaptation module employs learnable embeddings $\boldsymbol{E} \in \mathbb{R}^{N \times C}$, where $N$ represents predefined underwater conditions, to model degradation variations. By learning distinct embeddings, the module dynamically modulates features based on the observed scene. A per-pixel environmental descriptor is first computed by projecting feature maps $\boldsymbol{F}_c$ into the environmental embedding space:
\vspace{-8pt}
\begin{equation} 
\boldsymbol{E}_{adapted} = \sigma_{soft}(Linear(\boldsymbol{F}_c)) \otimes \boldsymbol{E},
\end{equation}
where $\sigma_{soft}$ is the Softmax function, ensuring each pixel receives a probabilistic weighting over environmental types. This allows the model to emphasize features relevant to specific conditions, such as light absorption, scattering, and turbidity. The computed priors then modulate feature representations through a weighted gating mechanism:

\vspace{-4pt}
\begin{equation} 
\boldsymbol{F}_e = \phi(\boldsymbol{F}_c \odot \sigma_{sig}(\boldsymbol{E}_{adapted})),
\end{equation}
where $\sigma_{sig}$ denotes the Sigmoid function and $\phi$ is the GELU activation. This formulation enhances relevant features while mitigating underwater degradations like light attenuation and color distortion. Finally, a zero-initialized up-projection step follows \cite{zhang2023adding} to stabilize early training while preserving environmental priors:

\vspace{-6pt}
\begin{equation} 
\boldsymbol{F}_a = \boldsymbol{F} + \mathcal{Z}(\boldsymbol{F}_e),
\label{equation13}
\end{equation}
where $\mathcal{Z}$ denotes the zero-initialized projection function. This ensures smooth adaptation without introducing excessive artifacts. The overall process enhances scene-aware feature modulation, improving robustness across diverse underwater conditions.


\begin{table*}[t]
\fontsize{9pt}{13pt}\selectfont
\centering
\begin{tabular}{p{3.5cm}@{\hspace{0.3cm}}|
>{\centering\arraybackslash}p{1.4cm}@{\hspace{0.3cm}}
>{\centering\arraybackslash}p{0.6cm}@{\hspace{0.3cm}}
>{\centering\arraybackslash}p{0.6cm}@{\hspace{0.3cm}}
>{\centering\arraybackslash}p{0.6cm}@{\hspace{0.3cm}}
>{\centering\arraybackslash}p{1.4cm}|@{\hspace{0.3cm}}
>{\centering\arraybackslash}p{2.2cm}@{\hspace{0.3cm}}
>{\centering\arraybackslash}p{0.6cm}@{\hspace{0.3cm}}
>{\centering\arraybackslash}p{0.6cm}@{\hspace{0.3cm}}
>{\centering\arraybackslash}p{0.6cm}@{\hspace{0.3cm}}
>{\centering\arraybackslash}p{1.4cm}@{\hspace{0.3cm}}}
\toprule\specialrule{0.1em}{0pt}{1pt}
\rowcolor{gray!15}
\multicolumn{11}{c}{\raisebox{1pt}{\rule{0pt}{7pt}\bf{Underwater Image Instance Segmentation (UIIS)}}}\\
\centering
\bf{Method}&\bf{Backbone}&\bf{mAP}&\bf{AP$_{50}$}&\bf{AP$_{75}$}&\bf{Params}&\bf{Backbone}&\bf{mAP}&\bf{AP$_{50}$}&\bf{AP$_{75}$}&\bf{Params}\\
\specialrule{0em}{1pt}{0pt}\hline\specialrule{0em}{1pt}{2pt}
Mask R-CNN \cite{he2017mask}&Swin-B&28.2&46.6&32.1&106.75 M&ConvNeXt V2-B&28.5&46.0&32.3&107.70 M\\
Cascade Mask R-CNN \cite{cai2018cascade}&Swin-B&29.4&48.0&32.7&139.79 M&ConvNeXt V2-B&28.2&45.2&32.4&140.74 M\\
Point Rend \cite{kirillov2020pointrend}&Swin-B&29.7&47.7&32.2&118.84 M&ConvNeXt V2-B&30.0&47.7&32.3&119.79 M\\
SOLOv2 \cite{wang2020solov2}&Swin-B&28.6&45.4&30.6&109.00 M&ConvNeXt V2-B&\textcolor{blue}{30.8}&47.7&33.9&109.95 M\\
Mask2Former \cite{cheng2022masked}&Swin-B&\textcolor{blue}{30.3}&45.6&32.4&106.75 M&ConvNeXt V2-B&25.1&38.9&26.7&107.70 M\\
WaterMask \cite{lian2023watermask}&Swin-B&30.1&\textcolor{blue}{49.0}&\textcolor{blue}{33.5}&110.40 M&ConvNeXt V2-B&30.1&\textcolor{blue}{48.3}&\textcolor{blue}{34.4}&111.35 M\\
USIS-SAM \cite{lian2024diving}&ViT-H&29.4&45.0&32.3&698.12 M&-&-&-&-&-\\
\bf{BARIS-ERA (Ours)}&Swin-B&\textcolor{red}{31.6}&\textcolor{red}{52.0}&\textcolor{red}{33.6}&114.44 M&ConvNeXt V2-B&\textcolor{red}{32.3}&\textcolor{red}{51.4}&\textcolor{red}{36.3}&112.46 M\\
\bottomrule\specialrule{0.1em}{0pt}{1pt}
\end{tabular}\vspace{-4pt}
\caption{Quantitative comparison with state-of-the-art methods on the UIIS dataset. USIS-SAM \cite{lian2024diving} uses a ViT-H backbone, while all other methods adopt Swin-B and ConvNeXt V2-B backbones. \textcolor{red}{Red} indicates the best performance, and \textcolor{blue}{blue} indicates the second-best.}
\vspace{-6pt}
\label{table1}
\end{table*}

\begin{table}[t]
\fontsize{9pt}{12pt}\selectfont
\centering
\begin{tabular}{p{2.7cm}@{\hspace{0.5cm}}
>{\centering\arraybackslash}p{1.6cm}@{\hspace{0.5cm}}
>{\centering\arraybackslash}p{0.4cm}@{\hspace{0.5cm}}
>{\centering\arraybackslash}p{0.4cm}@{\hspace{0.5cm}}
>{\centering\arraybackslash}p{0.4cm}@{\hspace{0.5cm}}
}
\toprule\specialrule{0.1em}{0pt}{1pt}
\rowcolor{gray!15}
\multicolumn{5}{c}{\raisebox{0pt}{\rule{0pt}{7pt}\bf{Underwater Salient Instance Segmentation (USIS10K)}}}\\
\centering
\multirow{2}{*}{\bf{Method}} & \multirow{2}{*}{\bf{Backbone}} & \multicolumn{3}{c}{\raisebox{0pt}{\rule{0pt}{7pt}\bf{Multi-Class}}} \\
&&\bf{mAP}&\bf{AP$_{50}$}&\bf{AP$_{75}$}\\
\specialrule{0em}{1pt}{-1pt}\hline\specialrule{0em}{1pt}{2pt}
WaterMask \cite{lian2023watermask}&ResNet-101&38.7&54.9&43.2\\
WaterMask \cite{lian2023watermask}&Swin-B&44.2&61.5&49.6\\
RSPrompter \cite{chen2024rsprompter}&ViT-H&40.2&55.3&44.8\\
USIS-SAM \cite{lian2024diving}&ViT-H&43.1&59.0&48.5\\
\bf{BARIS-ERA (Ours)}&Swin-B&\bf{47.3}&\bf{65.1}&\bf{53.7}\\
\bottomrule\specialrule{0.1em}{0pt}{1pt}
\end{tabular}\vspace{-4pt}
\caption{Quantitative comparisons with state-of-the-arts methods on the USIS10K datasets. BARIS-ERA follows the same hyperparameters and settings as in Table \ref{table1}. \textbf{Bold:} best.}
\vspace{-8pt}
\label{table2}
\end{table}

\subsection{Boundary-Aware Cross-Entropy Loss} \label{BACE_loss}
\vspace{-3pt}
Boundary-Aware Cross-Entropy (BACE) Loss enhances segmentation mask precision by leveraging range-null space decomposition, a fundamental concept in linear algebra widely applied in inverse problems \cite{wang2023gan, wang2022zero}. We observe that when applied to segmentation, this decomposition effectively preserves non-boundary structures while refining ambiguous edges, facilitating clearer and more accurate boundary representations.

\subsubsection{Range-Null Space Decomposition}
Given a transformation matrix $\boldsymbol{A} \in \mathbb{R}^{d \times D}$, its pseudo-inverse $\boldsymbol{A}^\dagger \in \mathbb{R}^{D \times d}$ satisfies:
\begin{equation}
    \boldsymbol{A} \boldsymbol{A}^\dagger \boldsymbol{A} = \boldsymbol{A}.
\end{equation}
Any vector $\boldsymbol{x} \in \mathbb{R}^{D}$ can be decomposed into range-space and null-space components:
\begin{equation}
    \boldsymbol{x} = \boldsymbol{A}^\dagger \boldsymbol{A} \boldsymbol{x} + (\boldsymbol{I} - \boldsymbol{A}^\dagger \boldsymbol{A})\boldsymbol{x}.
\end{equation}
The term $\boldsymbol{A}^\dagger \boldsymbol{A} \boldsymbol{x}$ projects $\boldsymbol{x}$ onto the range space of $\boldsymbol{A}$, preserving its essential structure. The term $(\boldsymbol{I} - \boldsymbol{A}^\dagger \boldsymbol{A}) \boldsymbol{x}$ projects $\boldsymbol{x}$ onto the null space of $\boldsymbol{A}$, capturing missing high-frequency details. This decomposition, originally used in inverse problems, allows range-space components to retain global structures while null-space components refine missing content.

\subsubsection{Application in Instance Segmentation}
For instance segmentation, we employ range-null space decomposition to refine mask predictions. Given the ground truth mask $\boldsymbol{M}_{\text{gt}}$ and the predicted mask $\boldsymbol{M}_\theta$, the refined prediction $\Gamma$ is formulated as:
\vspace{-2pt}
\begin{equation}
    \Gamma(\boldsymbol{M}_\theta, \boldsymbol{M}_{\text{gt}}) = \boldsymbol{A}^T \boldsymbol{A} \boldsymbol{M}_{\text{gt}} + (\boldsymbol{I} - \boldsymbol{A}^T \boldsymbol{A}) \boldsymbol{M}_\theta.
\end{equation}
The term $\boldsymbol{A}^T \boldsymbol{A} \boldsymbol{M}_{\text{gt}}$ ensures consistency with non-boundary regions by projecting the ground truth mask into the learned range space. The term $(\boldsymbol{I} - \boldsymbol{A}^T \boldsymbol{A}) \boldsymbol{M}_\theta$ refines the predicted mask by emphasizing details in the null space, enhancing boundary precision.

In implementation, $\boldsymbol{A}$ is a max-pooling operator extracting dominant structures, while $\boldsymbol{A}^T$ is a nearest-neighbor interpolation restoring boundary details lost during pooling. This decomposition effectively separates low-frequency global structures (range-space) from high-frequency boundary refinements (null-space), yielding sharper, more precise mask predictions.

\subsubsection{Final Loss Function}
The BACE Loss is formulated as:
\vspace{-8pt}
\begin{equation}
    \mathcal{L}_{BACE}(\boldsymbol{M}_\theta, \boldsymbol{M}_{gt}) = \frac{1}{N} \sum_{i=1}^{N} BCE(\boldsymbol{M}^i_{gt}, \Gamma(\boldsymbol{M}_\theta, \boldsymbol{M}_{gt})^i).
\end{equation}
The total optimization objective is:
\vspace{-2pt}
\begin{align}
    \mathcal{L}_{Total}(\boldsymbol{M}_\theta, \boldsymbol{M}_{gt}) &= \mathcal{L}_{CE}(\boldsymbol{M}_\theta, \boldsymbol{M}_{gt}) \notag \\
    &\quad + \lambda \cdot \mathcal{L}_{BACE}(\boldsymbol{M}_\theta, \boldsymbol{M}_{gt}),
\end{align}
where $\lambda = 1$ balances standard cross-entropy and BACE Loss. This formulation enhances global segmentation accuracy while refining boundary details, leading to more precise mask predictions.

%% file: sec/4_experiments.tex
\section{Experiments}
\vspace{-4pt}
We conduct extensive experiments to evaluate the effectiveness of BARIS-Decoder, ERA-tuning, and BACE Loss across multiple instance segmentation benchmarks. Section \ref{Implementation} details our experimental setup, including datasets, baseline models, and evaluation metrics. Section \ref{SOTA} presents comparisons with state-of-the-art methods, while Section \ref{Fine-Tuning} explores the efficiency and adaptability of ERA-tuning. In Section \ref{Ablation}, we conduct ablation studies to analyze the contributions of each module.

\subsection{Implementation Details}\label{Implementation}
\vspace{-3pt}
We evaluate our approach on the Underwater Image Instance Segmentation (UIIS) dataset \cite{lian2023watermask} and the Underwater Salient Instance Segmentation (USIS10K) dataset \cite{lian2024diving}. The UIIS dataset consists of 3,937 training images and 691 validation images, covering diverse underwater visibility conditions. USIS10K, a larger dataset, includes 10,632 images with more complex underwater environments.

We compare BARIS-ERA against leading instance segmentation frameworks, including Mask R-CNN \cite{he2017mask}, Cascade Mask R-CNN \cite{cai2018cascade}, PointRend \cite{kirillov2020pointrend}, SOLOv2 \cite{wang2020solov2}, Mask2Former \cite{cheng2022masked}, WaterMask \cite{lian2023watermask}, and USIS-SAM \cite{lian2024diving}. Additionally, we evaluate ERA-tuning against mainstream parameter-efficient tuning methods, including BitFit \cite{zaken2021bitfit, cai2020tinytl}, NormTuning \cite{giannou2023expressive}, PARTIAL-1 \cite{yosinski2014transferable}, Adapter \cite{houlsby2019parameter}, LoRA \cite{hu2021lora}, AdapterFormer \cite{chen2022adaptformer}, and MONA \cite{yin2023adapter}.

To ensure a fair comparison, all methods—except USIS-SAM, which employs a ViT-H backbone—use Swin Transformer \cite{liu2021swin} or ConvNeXt V2 \cite{liu2022convnet}, both pre-trained on ImageNet-22k \cite{deng2009imagenet}. The models are implemented in PyTorch \cite{paszke2017automatic} using OpenMMLab \cite{chen2019mmdetection}, and training is conducted on an Nvidia Titan RTX GPU. Hyperparameter details, including learning rate schedules and batch sizes, are provided in the supplementary material. For evaluation, we report mask AP \cite{lin2014microsoft} metrics, including mAP, AP$_{50}$, AP$_{75}$, AP$_{S}$, AP$_{M}$, and AP$_{L}$, ensuring comprehensive assessment across different IoU thresholds and object sizes.

%% file: sec/5_results.tex
\begin{figure}[t]
  \centering
  \includegraphics[width=1.0\linewidth]{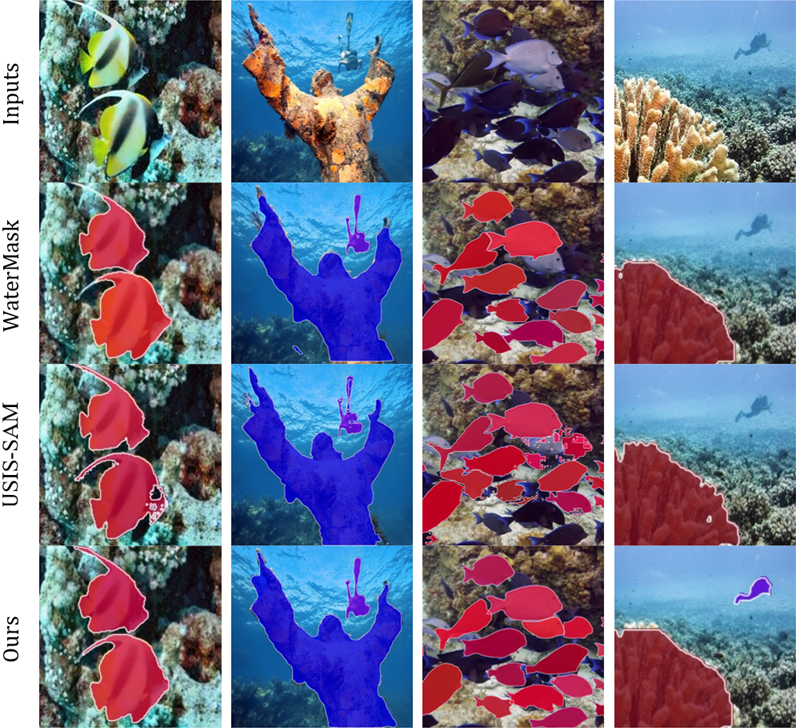}
  \vspace{-14pt}
  \caption{Qualitative comparison with state-of-the-art methods on the UIIS Dataset, using Swin Transformer (left two images) and ConvNeXt V2 (right two images) backbones.}
  \label{fig:fig6}
  \vspace{-14pt}
\end{figure}

\begin{table*}[t]
\fontsize{9pt}{10pt}\selectfont
\centering
\begin{tabular}{p{2.5cm}@{\hspace{0.5cm}}>{\centering\arraybackslash}p{1.1cm}@{\hspace{0.5cm}}>{\centering\arraybackslash}p{1.3cm}@{\hspace{0.5cm}}>{\centering\arraybackslash}p{1.4cm}@{\hspace{0.5cm}}c@{\hspace{0.5cm}} c@{\hspace{0.5cm}}c@{\hspace{0.5cm}}c@{\hspace{0.5cm}}c@{\hspace{0.5cm}}c@{\hspace{0.25cm}}}
\toprule\specialrule{0.1em}{0pt}{3pt}
\centering \raisebox{-6pt}{\bf{Method}}&\makebox[0pt][c]{\bf{Trained}}\makebox[4pt][c]{\raisebox{-11pt}{\bf{Params*}}}&\raisebox{-6pt}{\bf{\%}}&\bf{Extra Structure}& \raisebox{-6pt}{\bf{mAP}}&\raisebox{-6pt}{\bf{AP$_{50}$}}&\raisebox{-6pt}{\bf{AP$_{75}$}}&\raisebox{-6pt}{\bf{AP$_{S}$}}&\raisebox{-6pt}{\bf{AP$_{M}$}}&\raisebox{-6pt}{\bf{AP$_{L}$}}\\
\specialrule{0em}{1pt}{2pt}\hline\specialrule{0em}{1pt}{1pt}
\rowcolor{gray!15}
\multicolumn{10}{c}{\raisebox{1pt}{\rule{0pt}{8pt}\bf{Swin Transformer}}}\\
\specialrule{0em}{1pt}{1pt}\hline\specialrule{0em}{1pt}{1pt}
Full Fine-Tuning&\raggedleft 86.75 M&\raggedleft 100.00 \%&\ding{56}&28.2&46.6&\textcolor{blue}{32.1}&9.5&\textcolor{blue}{23.4}&39.6\\
BitFit \cite{zaken2021bitfit, cai2020tinytl}&\raggedleft 0.20 M&\raggedleft 0.23 \%&\ding{56}&26.0&46.8&25.5&8.4&21.9&36.9\\
NormTuning \cite{giannou2023expressive}&\raggedleft 0.06 M&\raggedleft 0.07 \%&\ding{56}&25.5&45.8&26.0&8.5&21.3&35.0\\
PARTIAL-1 \cite{yosinski2014transferable}&\raggedleft 12.60 M&\raggedleft 14.53 \%&\ding{56}&25.3&47.3&24.5&9.1&20.4&35.2\\
\hline\specialrule{0em}{1pt}{1pt}
Adapter \cite{houlsby2019parameter}&\raggedleft 3.11 M&\raggedleft 3.46 \%&\ding{51}&24.3&43.9&23.9&8.7&20.4&33.8\\
LoRA \cite{hu2021lora}&\raggedleft 3.08 M&\raggedleft 3.43 \%&\ding{51}&25.9&46.8&26.8&9.2&21.2&36.0\\
AdapterFormer \cite{chen2022adaptformer}&\raggedleft 1.55 M&\raggedleft 1.76 \%&\ding{51}&27.7&\textcolor{blue}{49.0}&29.6&9.5&22.6&38.8\\
MONA \cite{yin2023adapter}&\raggedleft 3.67 M&\raggedleft 4.06 \%&\ding{51}&\textcolor{blue}{28.9}&48.7&\textcolor{red}{32.5}&\textcolor{blue}{10.0}&22.5&\textcolor{blue}{41.4}\\
\hline\specialrule{0em}{2pt}{1pt}
\bf{ERA (Ours)}&\raggedleft 4.25 M&\raggedleft 4.67 \%&\ding{51}&\textcolor{red}{29.9}&\textcolor{red}{50.5}&\textcolor{red}{32.5}&\textcolor{red}{10.1}&\textcolor{red}{23.5}&\textcolor{red}{42.2}\\
\bottomrule\specialrule{0.1em}{0pt}{0pt}
\end{tabular}\vspace{-8pt}
\caption{Quantitative comparison with different fine-tuning methods on UIIS dataset using Swin Transformer backbones. \textcolor{red}{Red} indicates the best performance, and \textcolor{blue}{blue} indicates the second-best. * denotes the trainable parameters in backbones.}\vspace{-8pt}
\label{table3}
\end{table*}

\subsection{Comparison with State-of-the-Art Methods}\label{SOTA} \vspace{-4pt} 
We evaluate BARIS-ERA on the UIIS and USIS10K datasets, comparing its performance against leading instance segmentation methods. As shown in Table \ref{table1}, BARIS-ERA consistently outperforms prior methods on the UIIS dataset. With the Swin-B backbone, our method improves mAP by 3.4, 1.3, and 1.5 over Mask R-CNN \cite{he2017mask}, Mask2Former \cite{cheng2022masked}, and WaterMask \cite{lian2023watermask}, respectively. With ConvNeXt V2-B, it surpasses Mask R-CNN, Mask2Former, and SOLOv2 \cite{wang2020solov2} by 3.8, 7.2, and 1.5 mAP.

Table \ref{table2} further validates our method on the USIS10K dataset, where BARIS-ERA outperforms WaterMask by 3.1 mAP and USIS-SAM \cite{lian2024diving}, which employs a ViT-H backbone, by 4.2 mAP. These results confirm the effectiveness of our approach across diverse underwater segmentation scenarios.

Figure \ref{fig:fig6} qualitatively compares segmentation results. BARIS-ERA achieves more precise object boundaries, captures fine-grained details, and mitigates over-segmentation in dense regions. Compared to WaterMask and USIS-SAM, it better segments occluded objects and preserves structural integrity, even under challenging conditions like turbidity and lighting distortions. These results highlight the robustness of BARIS-ERA in real-world underwater applications.

\begin{figure}[t]
  \centering
  \includegraphics[width=1.0\linewidth]{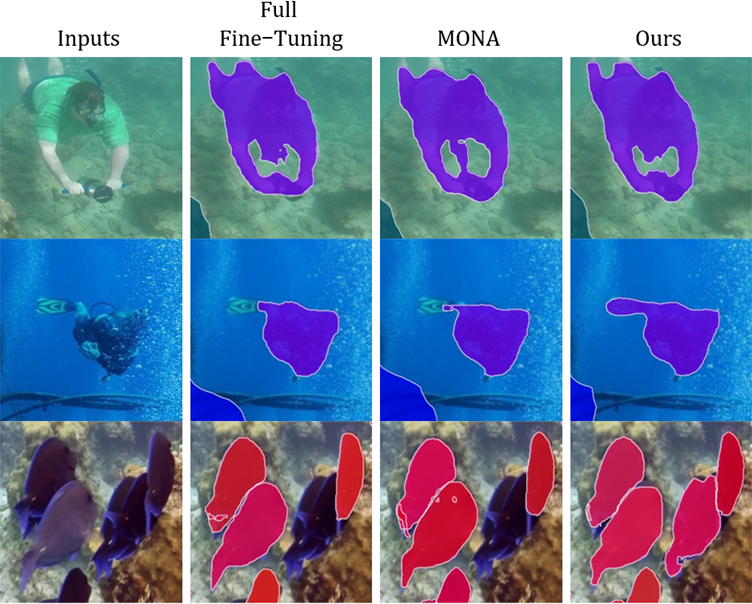}
  \caption{Qualitative comparison with different fine-tuning methods on the UIIS dataset. The first column is the original image. The second column shows full fine-tuning results. The third column displays MONA-tuning results. The fourth column shows the results of our ERA method.}
  \label{fig:fig7}
  \vspace{-8pt}
\end{figure}

\subsection{Comparison with Fine-Tuning Methods}\label{Fine-Tuning}
\vspace{-4pt} 
We evaluated ERA against various fine-tuning techniques using Swin Transformer backbones on the UIIS dataset. To ensure a fair comparison, we adjust the number of ERA parameters by modifying the compression ratio $\gamma$ so that its trainable parameter count closely matches that of MONA. This adjustment ensures that the observed improvements come from the effectiveness of the ERA rather than differences in the parameter budget, highlighting the efficiency of our approach. As shown in Table \ref{table3}, ERA achieves the highest mAP of 29.9, surpassing full fine-tuning by 1.7 mAP while using only 4.67\% of the trainable parameters. Compared to MONA \cite{yin2023adapter}, which achieves 28.9 mAP, ERA further improves performance by 1.0 mAP.

Figure \ref{fig:fig7} qualitatively compares ERA with full fine-tuning and MONA. ERA better preserves object boundaries and reduces segmentation errors, particularly in challenging underwater conditions with turbidity and lighting variations. Compared to other methods, ERA produces more complete segmentations and retains finer details, reinforcing its robustness in underwater instance segmentation.

\begin{table}[t]
\fontsize{8pt}{10pt}\selectfont
\centering
\begin{tabular}{p{2.2cm}@{\hspace{0.2cm}}c@{\hspace{0.1cm}}c@{\hspace{0.1cm}}c@{\hspace{0.1cm}} c@{\hspace{0.1cm}}c@{\hspace{0.1cm}}c@{\hspace{0.2cm}}c@{\hspace{0.1cm}}}
\toprule\specialrule{0em}{0pt}{1pt}
\centering \bf{Method}&\bf{mAP}&\bf{AP$_{50}$}&\bf{AP$_{75}$}&\bf{AP$_{S}$}&\bf{AP$_{M}$}&\bf{AP$_{L}$}&\bf{Params}\\
\specialrule{0em}{1pt}{-1pt}\hline\specialrule{0em}{1pt}{2pt}
Mask R-CNN&28.2&46.6&32.1&9.5&23.4&39.6&106.75 M\\
w/ BARIS-Decoder&30.0&49.2&32.1&9.5&23.7&42.8&105.14 M\\
w/ ERA&30.2&51.6&32.0&10.8&23.6&41.9&116.06 M\\
w/ BACE Loss&29.3&48.4&32.4&10.6&23.5&39.7&106.75 M\\
\bf{Full model (Ours)} &\bf{31.6}&\bf{52.0}&\bf{33.6}&\bf{10.7}&\bf{24.0}&\bf{45.0}&114.44 M\\
\bottomrule\specialrule{0em}{0pt}{2pt}
\end{tabular}\vspace{-8pt}
\caption{Effectiveness of each component. Swin-Transformer backbone and 1× training schedule is adopted. \textbf{Bold:} best.}
\vspace{-6pt}
\label{table4}
\end{table}

\subsection{Ablation Studies} \label{Ablation}
\vspace{-4pt} 
\textbf{Effectiveness of Each Component.}  
We analyze the contribution of each component in BARIS-ERA using the Swin Transformer backbone, as shown in Table \ref{table4}. The Mask R-CNN achieves an mAP of 28.2, serving as the baseline. Incorporating the BARIS-Decoder improves mAP to 30.0, enhancing feature boundaries, refining details, and strengthening multi-scale fusion. ERA-tuning further increases mAP to 30.2, demonstrating its effectiveness in mitigating underwater degradations and improving feature adaptability. BACE Loss boosts boundary refinement, achieving 29.3 mAP. The full model, integrating all components, attains the highest mAP of 31.6, confirming their complementary benefits for underwater instance segmentation.

\begin{table}[t]
\fontsize{8pt}{10pt}\selectfont
\centering
\begin{tabular}{p{2.99cm}@{\hspace{0.05cm}}c@{\hspace{0.05cm}}c@{\hspace{0.05cm}}c@{\hspace{0.05cm}}c@{\hspace{0.05cm}}c@{\hspace{0.05cm}}c@{\hspace{0.05cm}}c@{\hspace{0.05cm}}}
\toprule\specialrule{0em}{0pt}{1pt}
\centering \bf{Method}&\bf{mAP}&\bf{AP$_{50}$}&\bf{AP$_{75}$}&\bf{AP$_{S}$}&\bf{AP$_{M}$}&\bf{AP$_{L}$}&\bf{Params}\\
\specialrule{0em}{1pt}{-1pt}\hline\specialrule{0em}{1pt}{2pt}
Mask R-CNN&28.2&46.6&32.1&9.5&23.4&39.6&106.75 M\\
w/ RefineMask \cite{zhang2021refinemask}&29.7&47.8&\bf{32.8}&10.1&22.7&42.7&110.35 M\\
w/ WaterMask \cite{lian2023watermask}&29.3&46.7&32.5&\bf{10.5}&22.8&42.4&110.40 M\\
w/ BARIS-Decoder (Ours)&\bf{30.0}&\bf{49.2}&32.1&9.5&\bf{23.7}&\bf{42.8}&105.14 M\\
\bottomrule\specialrule{0em}{0pt}{2pt}
\end{tabular}\vspace{-8pt}
\caption{Effectiveness of refinement method. \textbf{Bold:} best.}
\label{table5}
\end{table}

\begin{table}[t]
\fontsize{8pt}{10pt}\selectfont
\centering
\begin{tabular}{p{3.5cm}@{\hspace{0.2cm}}c@{\hspace{0.1cm}}c@{\hspace{0.1cm}}c@{\hspace{0.1cm}} c@{\hspace{0.1cm}}c@{\hspace{0.1cm}}c@{\hspace{0.1cm}}}
\toprule\specialrule{0em}{0pt}{1pt}
\centering \bf{Method}&\bf{mAP}&\bf{AP$_{50}$}&\bf{AP$_{75}$}&\bf{AP$_S$}&\bf{AP$_M$}&\bf{AP$_L$}\\
\specialrule{0em}{1pt}{-1pt}\hline\specialrule{0em}{1pt}{2pt}
Cross Entropy Loss (CE)&28.2&46.6&32.1&9.5&23.4&39.6\\
CE + b-awareness Loss \cite{xu2023pidnet}&28.9&47.3&32.0&10.2&22.5&\bf{41.8}\\
CE + AB Loss \cite{wang2022active}&28.5&47.5&32.3&9.0&22.7&40.8\\
CE + BACE Loss (Ours)&\bf{29.3}&\bf{48.4}&\bf{32.4}&\bf{10.6}&\bf{23.5}&39.7\\
\bottomrule\specialrule{0em}{0pt}{2pt}
\end{tabular}\vspace{-8pt}
\caption{Effectiveness of boundary-aware loss. \textbf{Bold:} best.}
\label{table6}
\end{table}

\begin{table}[t]
\fontsize{8pt}{10pt}\selectfont
\centering
\begin{tabular}{p{1.8cm}@{\hspace{0.2cm}}c@{\hspace{0.1cm}}c@{\hspace{0.1cm}}c@{\hspace{0.1cm}} c@{\hspace{0.1cm}}c@{\hspace{0.1cm}}c@{\hspace{0.1cm}}c@{\hspace{0.1cm}}c@{\hspace{0.1cm}}c@{\hspace{0.1cm}}}
\toprule\specialrule{0em}{0pt}{1pt}
\centering \bf{\# Refine Block}&\bf{mAP}&\bf{AP$_{50}$}&\bf{AP$_{75}$}&\bf{AP$_{S}$}&\bf{AP$_{M}$}&\bf{AP$_{L}$}&\bf{Params}\\
\specialrule{0em}{1pt}{-1pt}\hline\specialrule{0em}{1pt}{2pt}
\centering 2&31.0&51.2&\bf{34.5}&10.2&\bf{24.6}&43.6&114.20 M\\
\centering 3&\bf{31.6}&\bf{52.0}&33.6&\bf{10.7}&24.0&\bf{45.0}&114.44 M\\
\centering 4&30.0&50.4&33.2&9.8&23.0&42.7&114.87 M\\
\centering 5&31.0&50.6&33.7&\bf{10.7}&24.1&44.3&115.69 M\\
\bottomrule\specialrule{0em}{0pt}{2pt}
\end{tabular}\vspace{-8pt}
\caption{The impact of the number of Refine Blocks. \textbf{Bold:} best.}
\vspace{-8pt} 
\label{table7}
\end{table}

\textbf{Effectiveness of Refinement Method.}  
To justify the design of BARIS-Decoder, we compare it with alternative refinement modules, including RefineMask \cite{zhang2021refinemask} and WaterMask \cite{lian2023watermask}, as shown in Table \ref{table5}. While both methods utilize multi-scale feature fusion, BARIS-Decoder incorporates a gated refinement mechanism that selectively enhances feature representation while preserving structural details. This leads to superior segmentation accuracy, achieving the highest mAP. The results validate the effectiveness of BARIS-Decoder in refining object boundaries and improving feature aggregation.

\textbf{Effectiveness of Different Boundary-Aware Loss.}  
Table \ref{table6} compares BACE Loss with other boundary-aware losses. Unlike b-awareness Loss from PIDNet, which applies weighted cross-entropy to emphasize edges, and Active Boundary Loss (ABL), which optimizes local boundary alignment, BACE Loss utilizes range-null space decomposition to refine boundary consistency while preserving global structure. This results in an mAP of 29.3, outperforming prior losses and demonstrating its effectiveness in challenging segmentation tasks.

\textbf{Impact of the Number of Refine Blocks.}  
We investigate the effect of varying the number of Refine Blocks on segmentation performance and computational efficiency, as shown in Table \ref{table7}. Increasing from two to three blocks improves mAP from 31.0 to 31.6, demonstrating the benefits of deeper feature refinement. However, further increasing to four or five blocks results in diminishing returns, with increased computational cost. Thus, we adopt three Refine Blocks as the optimal configuration, balancing segmentation quality and inference speed.

\vspace{6pt}
\textbf{The Impact of the Projection Ratio $\gamma$ in ERA.} We assessed the effect of the projection ratio $\gamma$ in ERA using Swin Transformer and ConvNeXt V2 backbones (see Table \ref{table8}). For Swin Transformer, $\gamma = 2$ achieved the highest mAP of 31.6, while $\gamma = 4$ balanced performance across multiple metrics. Higher ratios, such as $\gamma = 8$, led to declines in mAP. For ConvNeXt V2, $\gamma = 4$ yielded the best mAP of 32.3, with $\gamma = 2$ following closely behind. These results suggest that a lower $\gamma$ is optimal for Swin Transformer, while moderate values work best for ConvNeXt V2. We used the best configurations in all experiments, highlighting the importance of selecting an appropriate $\gamma$ for optimal ERA performance in underwater segmentation tasks.

\begin{table}[t]
\fontsize{8pt}{10pt}\selectfont
\centering
\begin{tabular}{p{1.5cm}@{\hspace{0.2cm}}c@{\hspace{0.1cm}}c@{\hspace{0.1cm}}c@{\hspace{0.1cm}} c@{\hspace{0.1cm}}c@{\hspace{0.1cm}}c@{\hspace{0.2cm}}c@{\hspace{0.1cm}}}
\toprule\specialrule{0em}{0pt}{1pt}
\centering \bf{Projection Ration ($\bm{\gamma}$)}&\raisebox{-4pt}{\bf{mAP}}&\raisebox{-4pt}{\bf{AP$_{50}$}}&\raisebox{-4pt}{\bf{AP$_{75}$}}&\raisebox{-4pt}{\bf{AP$_{S}$}}&\raisebox{-4pt}{\bf{AP$_{M}$}}&\raisebox{-4pt}{\bf{AP$_{L}$}}&\raisebox{-4pt}{\bf{Params}}\\
\hline
\rowcolor{gray!15}
\multicolumn{8}{c}{\raisebox{1pt}{\rule{0pt}{8pt}\bf{Swin Transformer}}}\\
\specialrule{0em}{1pt}{1pt}
\centering 2&\bf{31.6}&\bf{52.0}&\underline{33.6}&\bf{10.7}&\underline{24.0}&\bf{45.0}&114.44 M\\
\centering 4&\underline{30.6}&\underline{50.3}&\bf{34.5}&\underline{10.4}&\bf{24.2}&\underline{42.9}&109.38 M\\
\centering 8&29.3&48.7&32.9&10.0&23.9&41.5&107.18 M\\
\specialrule{0em}{1pt}{-1pt}\hline
\rowcolor{gray!15}
\multicolumn{8}{c}{\raisebox{1pt}{\rule{0pt}{8pt}\bf{ConvNeXt V2}}}\\
\specialrule{0em}{1pt}{1pt}
\centering 2&\underline{31.8}&\underline{51.0}&34.9&\underline{11.0}&\bf{24.0}&\underline{45.4}&120.05 M\\
\centering 4&\bf{32.3}&\bf{51.4}&\bf{36.3}&10.9&\underline{23.8}&\bf{45.7}&112.46 M\\
\centering 8&31.4&50.5&\underline{35.3}&\bf{11.3}&23.6&44.8&109.15 M\\
\bottomrule\specialrule{0em}{0pt}{2pt}
\end{tabular}\vspace{-8pt}
\caption{The impact of the projection ratio $\bm{\gamma}$ in ERA. Results are obtained using the Swin Transformer and ConvNeXt V2 backbones with a 1× training schedule. \textbf{Bold:} best, \underline{underline}: 2nd.}
\vspace{-8pt}
\label{table8}
\end{table}

%% file: sec/6_conclusion.tex
\vspace{-4pt}
\section{Conclusion}
\vspace{-6pt}
In this work, we introduce BARIS-Decoder and the Environmental Robust Adapter (ERA) to improve instance segmentation by refining boundary precision and adapting to environmental distortions. BARIS-Decoder enhances multi-scale feature processing through Multi-Stage Gated Refinement Network (MSGRN) and Depthwise Separable Upsampling (DSU), improving mask quality. ERA effectively counteracts degradation effects while significantly reducing trainable parameters, making adaptation more efficient. Additionally, Boundary-Aware Cross-Entropy (BACE) Loss further refines boundary consistency. Experimental results demonstrate that our BARIS-Decoder with ERA-tuning (BARIS-ERA) achieves state-of-the-art performance, surpassing prior methods in both segmentation accuracy and computational efficiency. 

Despite these advancements, challenges remain in extreme underwater conditions, such as severe turbidity and highly variable lighting, where object boundaries become difficult to delineate. Future work will address these limitations by improving robustness in degraded environments and extending evaluation to additional underwater datasets. Further optimizations will also focus on enhancing inference efficiency for real-time applications.

\vspace{10pt}


%% file: sec/X_suppl.tex
\clearpage
\setcounter{page}{1}
\twocolumn[
\begin{center}
    \vspace{1em}
    {\Large\bfseries BARIS: Boundary-Aware Refinement with Environmental Degradation Priors for Robust Underwater Instance Segmentation
    \section*{Supplementary Material}}
    \vspace{1em}
\end{center}
]

\section{Appendix} \label{sec}

\subsection{Training Setup}
\begin{itemize}
    \item \textbf{Swin Transformer Backbone:} We utilized a Mask R-CNN-based architecture with Swin Transformer as the backbone to leverage its powerful hierarchical representation and environmental adaptability features. Our setup includes the RefineMask module for multi-stage feature refinement and the ERA-tuning module to handle the domain shift inherent in underwater conditions. Key hyperparameters were set as follows: a base learning rate of 0.0001 was used with the AdamW optimizer, employing $(\beta_1, \beta_2) = (0.9, 0.999)$ for momentum parameters and a weight decay of 0.05 to prevent overfitting. A warmup phase was implemented with 1,000 iterations to gradually increase the learning rate, ensuring stable convergence. The model was trained for a total of 12 epochs, with a learning rate decay scheduled at epochs 8 and 11, following a step decay schedule to fine-tune performance in later stages.

    \item \textbf{ConvNeXt V2 Backbone:} Similarly, we used a Mask R-CNN-based architecture with ConvNeXt V2 as the backbone to explore its advantages in handling complex visual patterns common in underwater scenes. The core training configurations, including the optimizer, learning rate, warmup phase, and epoch schedule, mirrored those of the Swin Transformer backbone. We also incorporated environmental robustness features, tailoring ConvNeXt V2 with layer-wise decay to manage feature adaptation effectively. Specifically, a decay rate of 0.95 was applied over six layers, optimizing the balance between retaining pretrained knowledge and adapting to underwater specifics.
\end{itemize}
The configuration files included in our code repository provide an overview of additional setup details.

\subsection{Multi-Scale Feature Extraction Details} \vspace{-4pt}
The Multi-Scale Feature Extraction (MSFE) module enhances feature representation by capturing spatial information at multiple receptive fields while maintaining computational efficiency. Inspired by inception-style architectures, MSFE applies depthwise separable convolutions with varying kernel sizes ($3 \times 3$, $5 \times 5$, and $7 \times 7$), allowing the model to extract both fine-grained and large-scale contextual features. Additionally, max pooling followed by a $1 \times 1$ convolution is used to retain discriminative information while reducing spatial redundancy. The extracted features are aggregated to produce a refined feature representation.

\begin{figure}[t]
  \centering
  \includegraphics[width=1.0\linewidth]{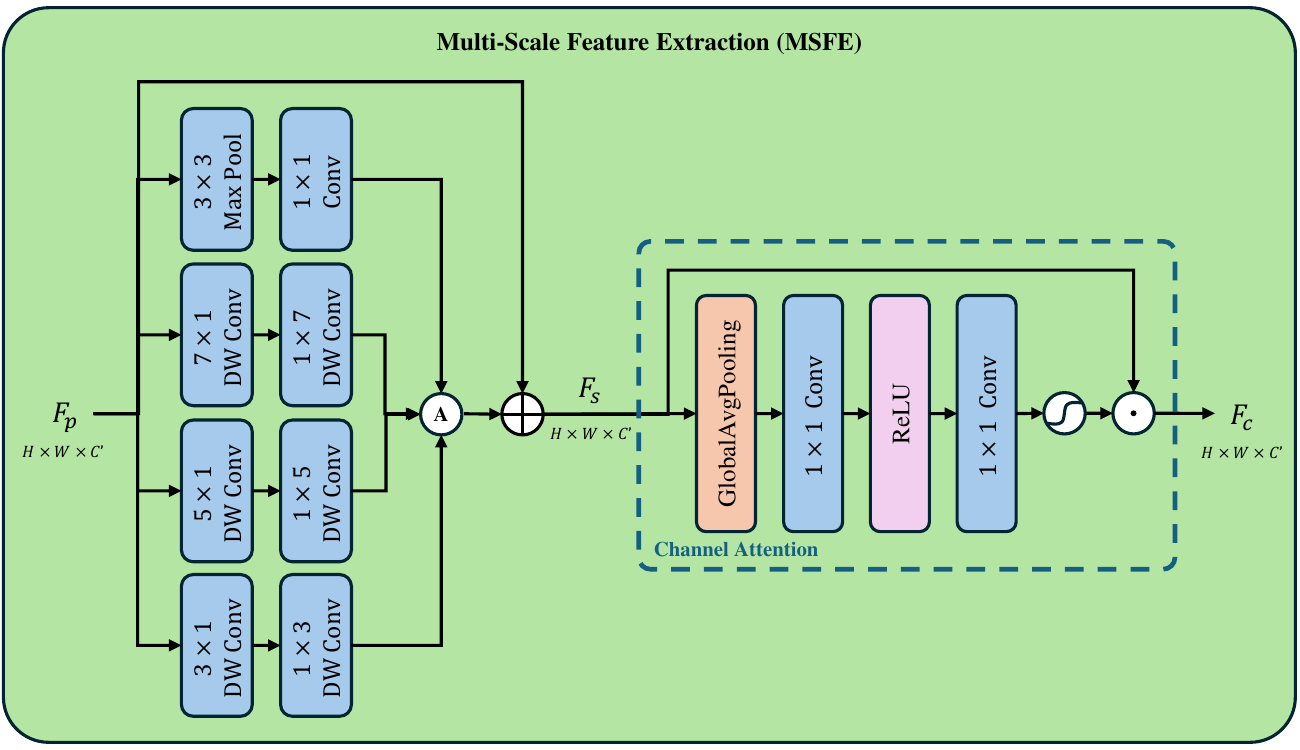}
  \caption{Architecture of the Multi-Scale Feature Extraction (MSFE) module. The MSFE integrates depthwise convolutions of various kernel sizes and channel attention to enhance feature extraction, addressing underwater challenges such as visibility degradation and color distortion.}
  \label{fig:fig10}
  \vspace{-10pt}
\end{figure}

To further improve feature discrimination, MSFE integrates a Channel Attention (CA) module. This mechanism applies global average pooling, followed by two $1 \times 1$ convolutions and a ReLU activation, to generate adaptive channel-wise attention weights. The refined features are then scaled accordingly, enhancing important features while suppressing less relevant ones. This design effectively preserves object boundaries and improves segmentation performance in complex scenes with overlapping objects and low-contrast regions. By leveraging multi-scale spatial feature extraction and adaptive channel weighting, MSFE achieves strong feature representation while maintaining computational efficiency. Figure \ref{fig:fig10} illustrates the architecture of MSFE, showcasing the combination of multi-scale depthwise convolutions, max pooling, and channel attention for robust feature learning.

\begin{figure*}[h]
\centering
\begin{lstlisting}[language=Python]
# Boundary-Aware Cross Entropy (BACE) Loss
def boundary_aware_cross_entropy(pred, label, scale, class_weight):
    # Downsample the prediction (A * pred) using max pooling
    A_pred = MaxPooling(pred, kernel_size=scale)
    
    # Upsample the result (A^T * A * pred) back to original size
    AtA_pred = Upsample(A_pred, scale_factor=scale)
    
    # Compute orthogonal projection (I - A^T * A) * pred
    ortho_project = pred - AtA_pred
    
    # Downsample the ground truth (A * label) using max pooling
    A_label = MaxPooling(label, kernel_size=scale)
    
    # Upsample the ground truth (A^T * A * label) back to original size
    AtA_label = Upsample(A_label, scale_factor=scale)
    
    # Compute parallel projection (A^T * A * label)
    parallel_project = AtA_label
    
    # Combine orthogonal and parallel projections for refined mask
    refined_pred = parallel_project + ortho_project
    
    # Compute the binary cross-entropy loss with logits
    loss = BinaryCrossEntropyWithLogits(refined_pred, label, weight=class_weight)
    
    return loss

# Example inputs: pred (prediction), label (ground truth)
# Set scale (e.g., scale=4), and class_weight if needed.
# Call boundary_aware_cross_entropy(pred, label, scale, class_weight)
\end{lstlisting}
\vspace{-8pt}
\end{figure*}

\begin{table}[t]
\fontsize{8pt}{10pt}\selectfont
\centering
\begin{tabular}{p{4.3cm}@{\hspace{0.5cm}}
>{\centering\arraybackslash}p{1.2cm}@{\hspace{0.5cm}}
>{\centering\arraybackslash}p{1cm}@{\hspace{0.5cm}}}
\toprule\specialrule{0.1em}{0pt}{1pt}
\rowcolor{gray!15}
\multicolumn{3}{c}{\raisebox{1pt}{\rule{0pt}{7pt}\bf{Swin Transformer}}}\\
\centering \bf{Method}&\bf{Params}&\bf{FPS}\\
\specialrule{0em}{1pt}{-1pt}\hline\specialrule{0em}{1pt}{2pt}
Mask R-CNN \cite{he2017mask}&106.75 M&8.325\\
Cascade Mask R-CNN \cite{cai2018cascade}&139.79 M&7.430\\
Point Rend \cite{kirillov2020pointrend}&118.84 M&7.430\\
SOLOv2 \cite{wang2020solov2}&109.00 M&6.775\\
Mask2Former \cite{cheng2022masked}&106.75 M&4.401\\
WaterMask \cite{lian2023watermask}&110.40 M&9.597\\
USIS-SAM \cite{lian2024diving}&698.12 M&2.750\\
\bf{BARIS-ERA (Ours)}&114.44 M&4.866\\
\bottomrule\specialrule{0.1em}{0pt}{1pt}
\end{tabular}
\vspace{-8pt}
\caption{Comparison of FPS and parameter efficiency among different instance segmentation methods using the Swin Transformer backbone.}
\vspace{-10pt}
\label{table9}
\end{table}

\begin{table*}[t]
\fontsize{9pt}{10pt}\selectfont
\centering
\begin{tabular}{p{2.5cm}@{\hspace{0.5cm}}>{\centering\arraybackslash}p{1.1cm}@{\hspace{0.5cm}}>{\centering\arraybackslash}p{1.3cm}@{\hspace{0.5cm}}>{\centering\arraybackslash}p{1.4cm}@{\hspace{0.5cm}}c@{\hspace{0.5cm}} c@{\hspace{0.5cm}}c@{\hspace{0.5cm}}c@{\hspace{0.5cm}}c@{\hspace{0.5cm}}c@{\hspace{0.25cm}}}
\toprule\specialrule{0.1em}{0pt}{3pt}
\centering \raisebox{-6pt}{\bf{Method}}&\makebox[0pt][c]{\bf{Trained}}\makebox[4pt][c]{\raisebox{-8pt}{\bf{Params*}}}&\raisebox{-6pt}{\bf{\%}}&\bf{Extra Structure}& \raisebox{-6pt}{\bf{mAP}}&\raisebox{-6pt}{\bf{AP$_{50}$}}&\raisebox{-6pt}{\bf{AP$_{75}$}}&\raisebox{-6pt}{\bf{AP$_{S}$}}&\raisebox{-6pt}{\bf{AP$_{M}$}}&\raisebox{-6pt}{\bf{AP$_{L}$}}\\
\specialrule{0em}{1pt}{2pt}\hline\specialrule{0em}{1pt}{1pt}
\rowcolor{gray!15}
\multicolumn{10}{c}{\raisebox{1pt}{\rule{0pt}{8pt}\bf{ConvNeXt V2}}}\\
\specialrule{0em}{1pt}{1pt}\hline\specialrule{0em}{1pt}{1pt}
Full Fine-Tuning&\raggedleft 87.69 M&\raggedleft 100.00 \%&\ding{56}&\textcolor{blue}{28.5}&46.0&\textcolor{blue}{32.3}&7.9&\textcolor{blue}{22.1}&\textcolor{blue}{40.9}\\
BitFit \cite{zaken2021bitfit, cai2020tinytl}&\raggedleft 0.13 M&\raggedleft 0.15 \%&\ding{56}&27.9&\textcolor{blue}{47.6}&29.9&\textcolor{blue}{9.8}&21.9&38.0\\
NormTuning \cite{giannou2023expressive}&\raggedleft 0.04 M&\raggedleft 0.05 \%&\ding{56}&26.5&47.1&28.0&9.4&21.2&36.6\\
PARTIAL-1 \cite{yosinski2014transferable}&\raggedleft 8.46 M&\raggedleft 9.64 \%&\ding{56}&26.0&46.6&27.1&8.0&21.4&36.2\\
\hline\specialrule{0em}{1pt}{1pt}
VPT \cite{bahng2022exploring, jia2022visual}&\raggedleft 0.20 M&\raggedleft 0.23 \%&\ding{51}&26.8&47.2&28.0&\textcolor{blue}{9.8}&20.8&36.5\\
Conv-Adapter \cite{chen2024conv}&\raggedleft 2.36 M&\raggedleft 2.63 \%&\ding{51}&24.4&43.7&25.5&8.9&19.0&34.9\\
\hline\specialrule{0em}{2pt}{1pt}
\bf{ERA (Ours)}&\raggedleft 1.54 M&\raggedleft 1.72 \%&\ding{51}&\textcolor{red}{29.9}&\textcolor{red}{50.2}&\textcolor{red}{33.2}&\textcolor{red}{11.3}&\textcolor{red}{22.9}&\textcolor{red}{41.3}\\
\bottomrule\specialrule{0.1em}{0pt}{0pt}
\end{tabular}\vspace{-4pt}
\caption{Quantitative comparison with different fine-tuning methods on UIIS dataset using ConvNeXt V2 backbones. \textcolor{red}{Red} indicates the best performance, and \textcolor{blue}{blue} indicates the second-best. * denotes the trainable parameters in backbones.}
\vspace{-8pt}
\label{table10}
\end{table*}

\vspace{-4pt}
\subsection{PyTorch-Like Code Implementation} \vspace{-8pt}
We provide a PyTorch-like implementation of Boundary-Aware Cross-Entropy (BACE) Loss, illustrating how range-null space decomposition enhances segmentation accuracy, particularly at object boundaries. This implementation projects predictions onto range-space and null-space components, refining object contours while preserving structural consistency. In the implementation, we first apply max pooling to downsample both predictions and ground truth masks, extracting dominant structures and reducing high-frequency noise. This is followed by nearest-neighbor interpolation to restore spatial resolution. The range-space component ensures consistency with non-boundary regions, while the null-space component captures finer details, correcting boundary misalignment. The final mask is computed by combining these components and applying Binary Cross-Entropy (BCE) Loss for segmentation supervision. The BACE Loss integrates seamlessly into modern segmentation pipelines with minimal computational overhead. Unlike standard loss functions, it explicitly refines boundary features, improving segmentation accuracy in complex scenarios. Its flexibility allows it to be used across different segmentation tasks with customizable linear operators $\boldsymbol{A}$, such as blurring or inpainting operators in inverse problems. Additionally, the scaling parameter in the implementation determines the downsampling factor, providing adaptability for different dataset resolutions and object complexities. Researchers and practitioners can easily incorporate this method into existing frameworks to enhance segmentation precision, particularly for tasks requiring fine-grained boundary refinement.

\subsection{Computational Efficiency Analysis} 
We provide a comparison of frames per second (FPS) to evaluate the computational efficiency of BARIS-ERA relative to baseline methods. Table \ref{table9} reports the FPS and parameter count for models using Swin Transformer backbones. While BARIS-ERA achieves state-of-the-art segmentation performance, it maintains competitive inference speed. Compared to standard Mask R-CNN, our method introduces a moderate computational overhead due to multi-scale refinement and adapter-based tuning. However, BARIS-ERA remains significantly more efficient than USIS-SAM, which employs a ViT-H backbone, leading to substantially higher computational costs. The trade-off between accuracy and efficiency underscores the suitability of BARIS-ERA for practical applications, balancing segmentation precision with feasible real-time performance.

\subsection{Additional Fine-Tuning Comparisons} \vspace{-4pt}
To complement the results in Table \ref{table3}, which compare fine-tuning methods on Swin Transformer, we provide additional results for ConvNeXt V2 backbones in Table \ref{table10}. This comparison follows the same experimental setup, ensuring that parameter efficiency and segmentation performance are fairly evaluated across different architectures. As shown in Table \ref{table10}, ERA achieves the highest mAP of 29.9, surpassing full fine-tuning by 1.4 mAP while requiring only 1.72\% of the trainable parameters. The results reinforce the effectiveness of ERA across different model architectures, demonstrating its ability to efficiently adapt to varying feature representations while maintaining strong segmentation performance. These findings further validate ERA as an efficient alternative to traditional full fine-tuning, significantly reducing computational overhead while maintaining state-of-the-art segmentation performance across different network backbones.

\begin{figure}[t]
  \centering
  \includegraphics[width=1.0\linewidth]{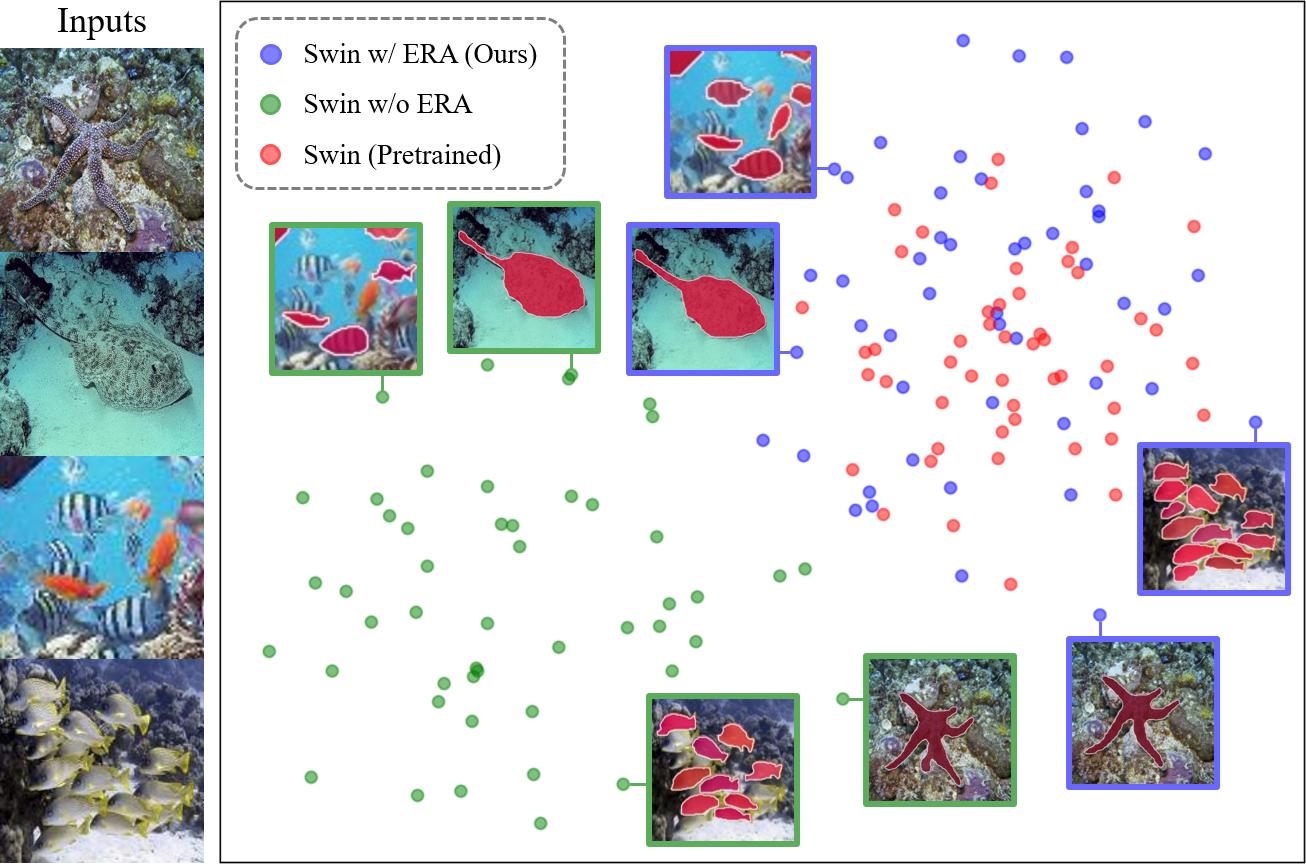}
  \vspace{-16pt}
  \caption{The t-SNE visualization of feature distributions illustrating the effectiveness of ERA in aligning underwater features with terrestrial distributions. "Swin w/ ERA (Ours)" (blue points) shows significant overlap with "Swin (Pretrained)," (red points) bridging the gap between underwater and land-based environments, while "Swin w/o ERA" (green points) remains distinct due to underwater-specific degradations.}
  \label{fig:fig8}
  \vspace{-6pt}
\end{figure}

\subsection{Knowledge Transfer of ERA} 
The purpose  of ERA is to adapt underwater image features by learning priors of various underwater degradations, allowing pretrained models on land-based data to process underwater imagery effectively. To evaluate the transferability of ERA, we present t-SNE visualizations in Figure \ref{fig:fig8}. The figure shows that "Swin w/o ERA" (green points), which uses full fine-tuning, captures underwater-specific features with distributions affected by underwater degradation (e.g., color distortions, low visibility). In contrast, "Swin (Pretrained)" (red points) retains ImageNet \cite{deng2009imagenet} features suited for terrestrial environments, demonstrating a distinct distribution. However, "Swin w/ ERA (Ours)" (blue points) achieves significant overlap with "Swin (Pretrained)," (red points) illustrating the effectiveness of the proposed ERA in dynamically adapting underwater features to align with terrestrial feature distributions by mitigating underwater degradation effects. This alignment is crucial for stabilizing training and capturing robust features in challenging underwater conditions. These results highlight the capability of ERA to adapt models for underwater segmentation, effectively bridging the gap between underwater and land-based visual characteristics.


\begin{figure}[t]
  \centering
  \includegraphics[width=1.0\linewidth]{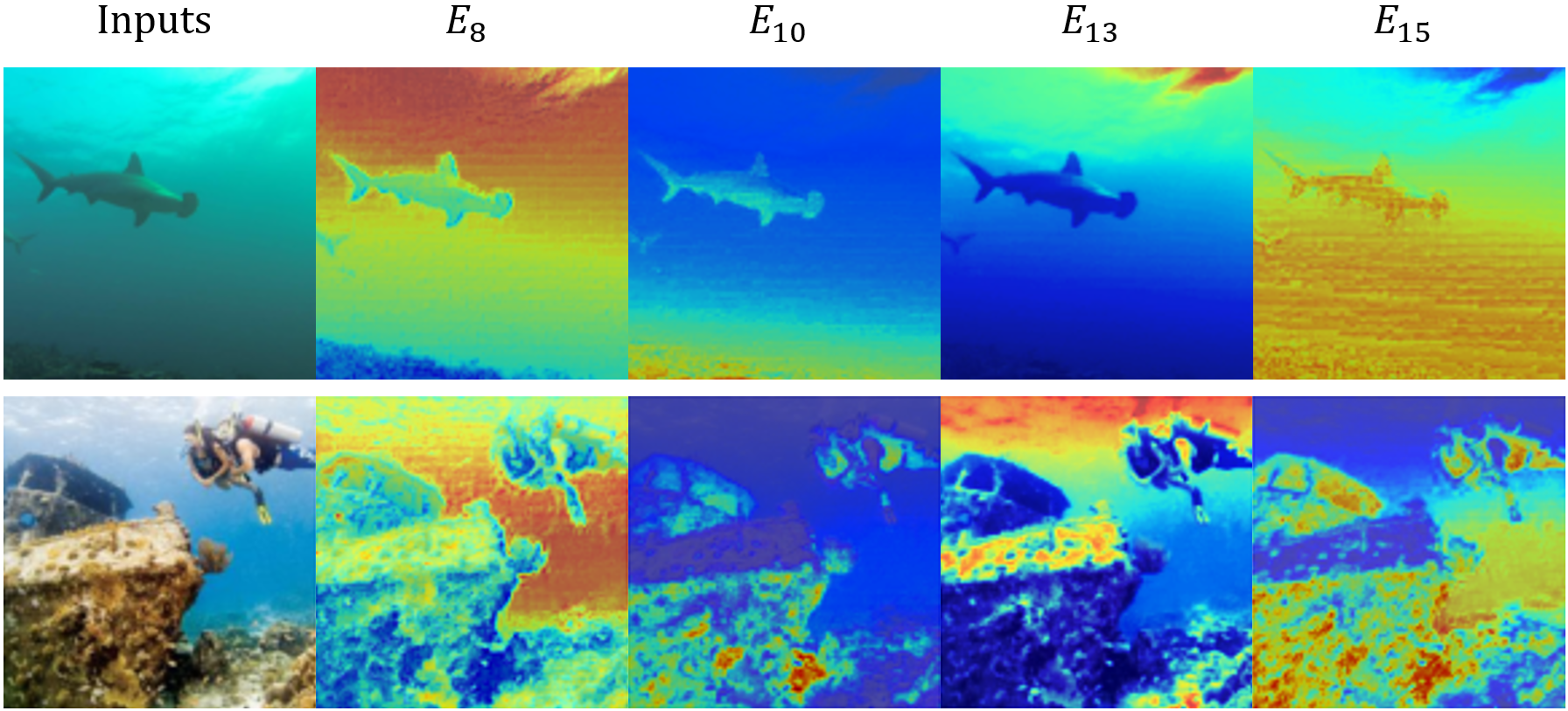}
  \vspace{-16pt}
  \caption{Visualization of learnable environmental degradation prior embeddings showing the effectiveness of our adaptation mechanism in addressing underwater degradations. Different embeddings complement each other in adapting to diverse underwater conditions.}
  \label{fig:fig9}
\end{figure}

\begin{table}[t]
\fontsize{8pt}{10pt}\selectfont
\centering
\begin{tabular}{p{1.8cm}@{\hspace{0.2cm}}c@{\hspace{0.1cm}}c@{\hspace{0.1cm}}c@{\hspace{0.1cm}} c@{\hspace{0.1cm}}c@{\hspace{0.1cm}}c@{\hspace{0.1cm}}}
\toprule\specialrule{0em}{0pt}{1pt}
\centering \bf{\# Environment Embeddings}&\raisebox{-4pt}{\bf{mAP}}&\raisebox{-4pt}{\bf{AP$_{50}$}}&\raisebox{-4pt}{\bf{AP$_{75}$}}&\raisebox{-4pt}{\bf{AP$_{S}$}}&\raisebox{-4pt}{\bf{AP$_{M}$}}&\raisebox{-4pt}{\bf{AP$_{L}$}}\\
\specialrule{0em}{1pt}{-1pt}\hline\specialrule{0em}{1pt}{2pt}
\centering 4&\underline{30.9}&50.7&33.0&\bf{10.8}&24.4&43.5\\
\centering 8&30.6&50.8&\bf{33.8}&10.3&\underline{24.5}&43.0\\
\centering 16&\bf{31.6}&\bf{52.0}&\underline{33.6}&\underline{10.7}&24.0&\bf{45.0}\\
\centering 32&30.6&\underline{51.1}&33.5&10.6&\bf{24.9}&42.4\\
\bottomrule\specialrule{0em}{0pt}{2pt}
\end{tabular}\vspace{-8pt}
\caption{The impact of the number of learnable environment embeddings. Evaluation is conducted using the Swin Transformer backbone with a 1× training schedule. \textbf{Bold:} best, \underline{underline}: 2nd.}
\vspace{-15pt}
\label{table11}
\end{table}

\subsection{Learnable Environment Embeddings}
We evaluated the effect of varying the number of learnable environmental embeddings, testing configurations with 4, 8, 16, and 32 embeddings. As shown in Table \ref{table11}, the 16-embedding configuration achieved the highest mAP of 31.6 and the best AP$_{50}$ of 52.0, indicating strong accuracy. The 4-embedding setup yielded an mAP of 30.9, while 8 embeddings attained the highest AP$_{75}$ of 33.8 with a competitive mAP of 30.6. The 32-embedding configuration slightly underperformed with an mAP of 30.6. These results suggest that 16 embeddings strike the optimal balance for accuracy under varying underwater conditions. Figure \ref{fig:fig9} visualizes representative learnable environmental degradation prior embeddings ($\boldsymbol{E}_8$, $\boldsymbol{E}_{10}$, $\boldsymbol{E}_{13}$, $\boldsymbol{E}_{15}$), showing their complementary roles in mitigating challenges such as turbidity and reduced visibility, enabling adaptation to diverse underwater environments.
